\documentclass[runningheads]{llncs}

\usepackage{eccv}

\usepackage{eccvabbrv}

\usepackage{graphicx}
\usepackage{amsmath}
\usepackage{amssymb}
\usepackage{booktabs}

\usepackage{comment}
\usepackage{cite}
\usepackage{amsmath,amssymb} 
\usepackage{color}
\usepackage{varwidth}
\usepackage{xcolor}

\usepackage{verbatim}
\usepackage{multirow}
\usepackage{adjustbox}
\usepackage{array}
\usepackage{bbm}
\usepackage{mathtools}
\usepackage{pifont}
\newcommand{\cmark}{\ding{51}}
\newcommand{\xmark}{\ding{55}}
\usepackage{scalerel}
\usepackage{listings}
\newcolumntype{x}[1]{>{\centering\let\newline\\\arraybackslash\hspace{0pt}}p{#1}}

\usepackage[accsupp]{axessibility}  

\usepackage[colorlinks = true,
            linkcolor = blue,
            urlcolor  = red,
            citecolor = eccvblue,
            anchorcolor = blue]{hyperref}

\usepackage{orcidlink}

\begin{document}

\title{Scaling Up Personalized Image Aesthetic Assessment via Task Vector Customization} 

\titlerunning{Scaling Up Personalized Image Aesthetic Assessment}

\author{Jooyeol Yun\orcidlink{0000-0001-7853-7703} \and
Jaegul Choo\orcidlink{0000-0003-1071-4835}
}
\authorrunning{JY.~Yun and J.~Choo}

\institute{Korea Advanced Institute of Science and Technology (KAIST)\\
\email{blizzard072@kaist.ac.kr} \,\, \email{jchoo@kaist.ac.kr}
}

\maketitle

\vspace{-0.5cm}
\begin{abstract}
The task of personalized image aesthetic assessment seeks to tailor aesthetic score prediction models to match individual preferences with just a few user-provided inputs. 
However, the scalability and generalization capabilities of current approaches are considerably restricted by their reliance on an expensive curated database.
To overcome this long-standing scalability challenge, we present a unique approach that leverages readily available databases for general image aesthetic assessment and image quality assessment. 
Specifically, we view each database as a distinct image score regression task that exhibits varying degrees of personalization potential. 
By determining optimal combinations of task vectors, known to represent specific traits of each database, we successfully create personalized models for individuals. 
This approach of integrating multiple models allows us to harness a substantial amount of data. 
Our extensive experiments demonstrate the effectiveness of our approach in generalizing to previously unseen domains---a challenge previous approaches have struggled to achieve---making it highly applicable to real-world scenarios. 
Our novel approach significantly advances the field by offering scalable solutions for personalized aesthetic assessment and establishing high standards for future research. 
\footnote{\url{https://yeolj00.github.io/personal-projects/personalized-aesthetics/}}
\vspace{-0.2cm}
\end{abstract}

\section{Introduction}
Personalized image aesthetic assessment (PIAA) is an emerging field dedicated to developing aesthetic score prediction models that closely match an individual's aesthetic preferences based on a small set of user-provided samples. 
While they function as simple models that assign scores based on aesthetic quality, PIAA models enhance user experience in digital environments by personalizing tasks such as managing photo albums~\cite{karlsson2014mobile}, curating web-scale databases like LAION-Aesthetics~\cite{laion}, and even guiding generative models such as Stable Diffusion~\cite{ldm} to create images that align with individual tastes~\cite{wallace2023end}. 

In recent years, there has been a notable trend in PIAA~\cite{blgpiaa, tapppiaa, mtcl} that involves the integration of meta-learning techniques, as personalization is inherently related to few-shot learning. 
Specifically, these approaches entail pre-training a model on a database consisting of image collections from diverse individuals. 
However, these strategies encounter significant challenges in terms of scalability and generalization to unseen domains, primarily due to the prohibitive cost associated with data collection. 

While meta-learning typically demands an extensive number of meta-training tasks, often exceeding 10,000, to achieve high performance~\cite{al2021data}, currently available PIAA databases~\cite{flickeraes, aadb} offer a limited number of tasks fewer than 200. 
Even with the recently collected PIAA database~\cite{para}, which contains rich attributes from each annotator, the number of training tasks remains below 500. 
In the context of PIAA, a single meta-training task corresponds to learning the preferences of a single individual. 
Scaling such a database would require obtaining a vast number of images and corresponding personalized aesthetic scores from numerous individuals, which poses severe limitations in the scalability and generalization of existing approaches. 


\begin{figure}[t]
    \centering
    \includegraphics[width=0.7\linewidth]{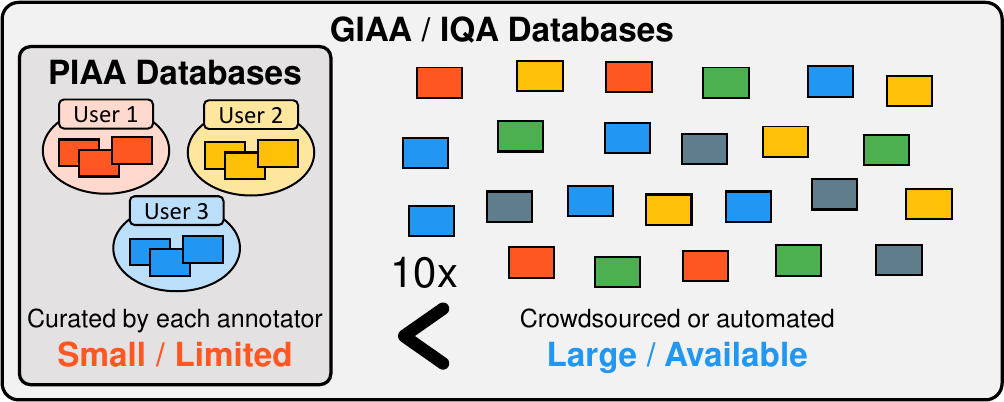}
    \vspace{-0.2cm}
    \caption{Comparison between PIAA and other image regression databases. Our approach has the flexibility to leverage any image regression databases. }
    \label{fig:intro-database-diff}
    \vspace{-0.6cm}
\end{figure}

In this paper, we unveil a data-rich approach for personalizing aesthetic assessment models by leveraging existing general image aesthetic assessment (GIAA) and image quality assessment (IQA) databases. 
In contrast to prior work~\cite{blgpiaa, piaasoa, paiaa, tapppiaa, mtcl}, our approach is not limited to training on databases that track individual annotators, as depicted in \Cref{fig:intro-database-diff}. 
Instead, we have the flexibility to utilize multiple image score regression databases, which are readily available. 
By employing broader resources, we overcome the scalability and generalization challenges that have \emph{long been unaddressed} in this field.

\begin{figure}[h]
    \centering
    \vspace{-0.6cm}
    \includegraphics[width=0.75\linewidth]{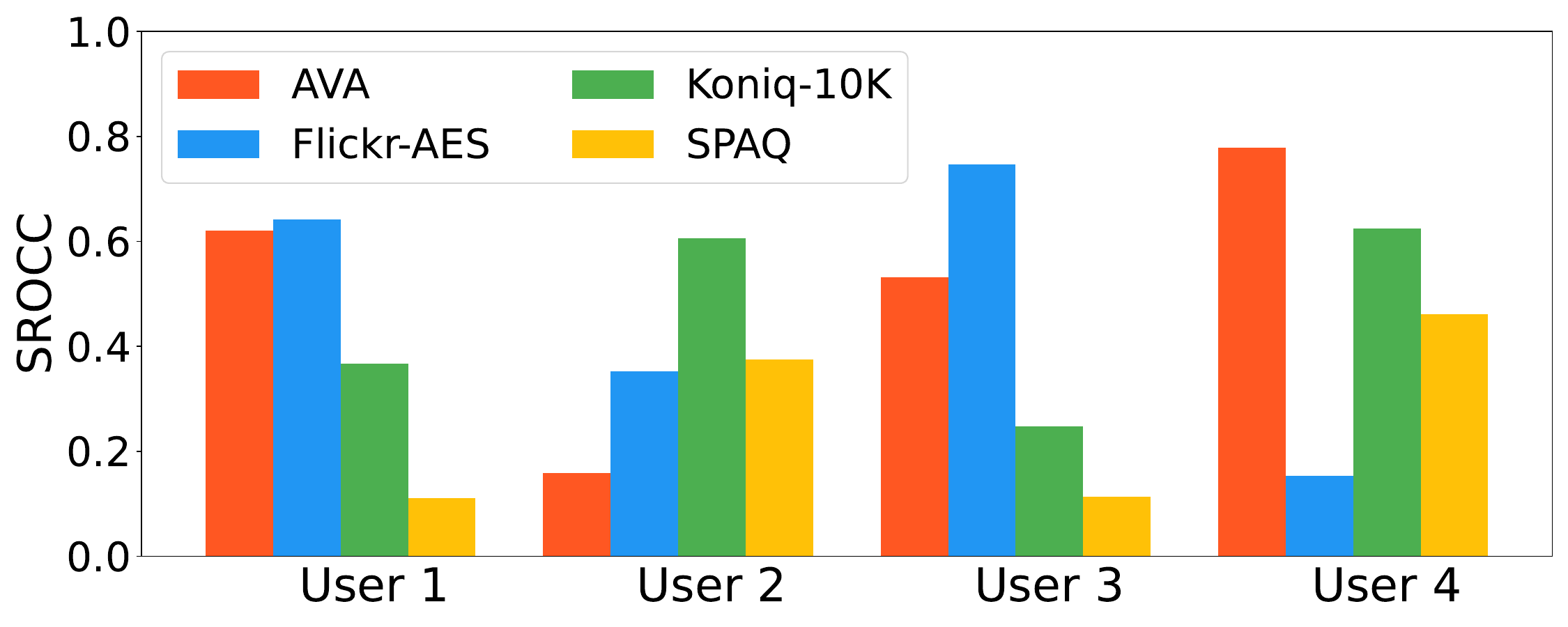}
    \vspace{-0.4cm}
    \caption{Zero-shot personalization performance for four users in the Flickr-AES database~\cite{flickeraes}. Each colored bar refers to the personalization performance of a model trained on a specific database. }
    \label{fig:intro-motivation}
    \vspace{-0.7cm}
\end{figure}

The key idea is to view each GIAA and IQA data as a unique image score regression task. 
This perspective arises from the distinct characteristics and tendencies of each database due to different data collection procedures and the population from which they originate. 
As visualized in \Cref{fig:intro-motivation}, image score regression models trained on four different GIAA~\cite{ava, flickeraes} and IQA~\cite{koniq10k, spaq} databases exhibit varying levels of personalization for users within the PIAA database. 
For instance, User~3's aesthetic preferences closely align with the score regression model trained on the Flickr-AES~\cite{flickeraes} while diverging from the model trained on the SPAQ~\cite{spaq} database. 
In other words, even without any personalized training, we find that general image assessment models inherently possess different degrees of personalization capabilities. 

In light of these insights, we extend an emerging approach known as task arithmetic~\cite{taskvector}, which was originally devised for developing multi-task models.
This method involves directly adding or subtracting the weights of models fine-tuned on specific tasks. 
In our context, task vectors from diverse image assessment models encapsulate distinct features and behaviors related to specific tasks, such as recognizing different aspects of image quality or aesthetics. 
By carefully combining task vectors, we explore the potential to selectively amplify or refine the capabilities of a pre-trained model. 

However, there has been limited research on determining the optimal adjustments necessary to achieve a certain behavior. 
Thus, we present a personalization approach by introducing \emph{trainable coefficients} for each task vector derived from models trained on distinct GIAA and IQA databases, thereby determining the weights for combining these vectors. 
Once the coefficients are trained with user-provided inputs, the weighted sum of task vectors creates a model tailored to the user's aesthetic preferences. 
Since task vectors obtained from large databases already capture preferences across various themes and contents, we find that training only the coefficients is sufficient for precise personalization, making our approach \emph{highly parameter-efficient}. 
To the best of our knowledge, our approach marks the first to customize task vectors for specific model behaviors that were previously unattainable. 

By carefully customizing task vectors, we incorporate large databases into the training process, granting our approach a strong generalization capability for unseen domains, a scenario often encountered in real-world applications. 
Furthermore, we find that our approach is also practical for learning a user's aesthetic preference with only a few samples, as our approach of merging task vectors guides the model to make \emph{well-informed updates} and prevents overfitting.
We provide extensive experiments demonstrating the efficacy of our approach for personalization in diverse scenarios.

Our contributions are threefold:
\vspace{-0.2cm}
\begin{itemize}
    \item[\textbullet] We introduce a novel data-rich approach for PIAA that tackles the long-standing scalability challenge, eliminating the dependency on expensive, manually-curated databases that have hindered progress in this field. 
    \vspace{0.1cm}
    \item[\textbullet] Demonstrating exceptional effectiveness in cross-database evaluations, our method showcases robust generalization capabilities that surpass those of existing approaches, setting high standards for future work. 
    \vspace{0.1cm}
    \item[\textbullet] By learning optimal combinations of task vectors, we present a parameter-efficient approach for precise personalization, leveraging the comprehensive information embedded in these vectors. 
\end{itemize}


\section{Related Work}
\vspace{-0.1cm}

\subsection{Personalized Image Aesthetic Assessment}
Personalized image aesthetic assessment (PIAA)~\cite{blgpiaa, paiaa, piaasoa, ugpiaa, usar, mtcl, tapppiaa, impiaa, para, pass, tcmlpiaa} aims to learn the personal aesthetic preference of individual users based on the image-score pairs from one's image collection. 
Since collecting a large number of samples from a single user is impractical, PIAA is closely linked to few-shot learning methods. 
One line of work~\cite{paiaa, piaasoa, para} incorporates additional attributes (\eg, personality traits, age, and gender) to facilitate user-specific aesthetic predictions.
These approaches rely on the correlation between user attributes and aesthetic preferences, which may not always be clearly defined. 
Consequently, many high-performing approaches for PIAA~\cite{blgpiaa, tapppiaa, mtcl} leverage meta-learning techniques to train models that can be easily fine-tuned to accommodate the aesthetic preferences of new users with only a limited number of samples. 

Nevertheless, applying meta-learning techniques to PIAA encounters scalability challenges owing to the limited number of tasks available during the meta-learning phase. 
In traditional meta-learning, achieving a high performance typically requires a substantial number of meta-training tasks~\cite{al2021data}, often exceeding 10,000. 
In PIAA databases~\cite{flickeraes, aadb}, the number of training tasks ranges from 100 to fewer than 500, which can impose limitations on the performance. 
Expanding the task pool in PIAA is prohibitively expensive since each task corresponds to the image collection of a single user, posing significant scalability challenges that has long been unaddressed. 
We tackle this issue in previous PIAA approaches and present a scalable fine-tuning method that leverages existing general image aesthetic assessment (GIAA) and image quality assessment (IQA) databases.

\vspace{-0.1cm}
\subsection{Task Vectors and Task Arithmetic}
Ilharco~\etal~\cite{taskvector} presented a new paradigm for creating multi-task models, utilizing the parameters of multiple fine-tuned models. 
They define the \emph{task vector} $\tau$ as the difference between the fine-tuned and pre-trained weights,
\begin{equation*}
    \tau = \theta_{\text{ft}} - \theta_{\text{pre}},
\end{equation*}
where $\theta_{\text{ft}}$ and $\theta_{\text{pre}}$ represent the weights of the fine-tuned and pre-trained model, respectively.
They demonstrate that adding and subtracting task vectors to pre-trained models can encourage or suppress the ability to perform specific tasks. 
For example, negating the task vector derived from a language model fine-tuned on toxic data reduces the frequency of generating toxic text. 
However, it remains unclear what the optimal adjustments are for achieving a specific behavior.

In our paper, we investigate methods to determine necessary adjustments to achieve a certain behavior defined by a few user-provided samples. 
By introducing learnable coefficients to task vectors, we effectively customize models to closely match individual preferences.
We also demonstrate that finding the optimal coefficients is possible with only a few samples, as the task vectors encompass comprehensive knowledge derived from large databases. 

\begin{figure*}[t]
    \centering
    \includegraphics[width=\textwidth]{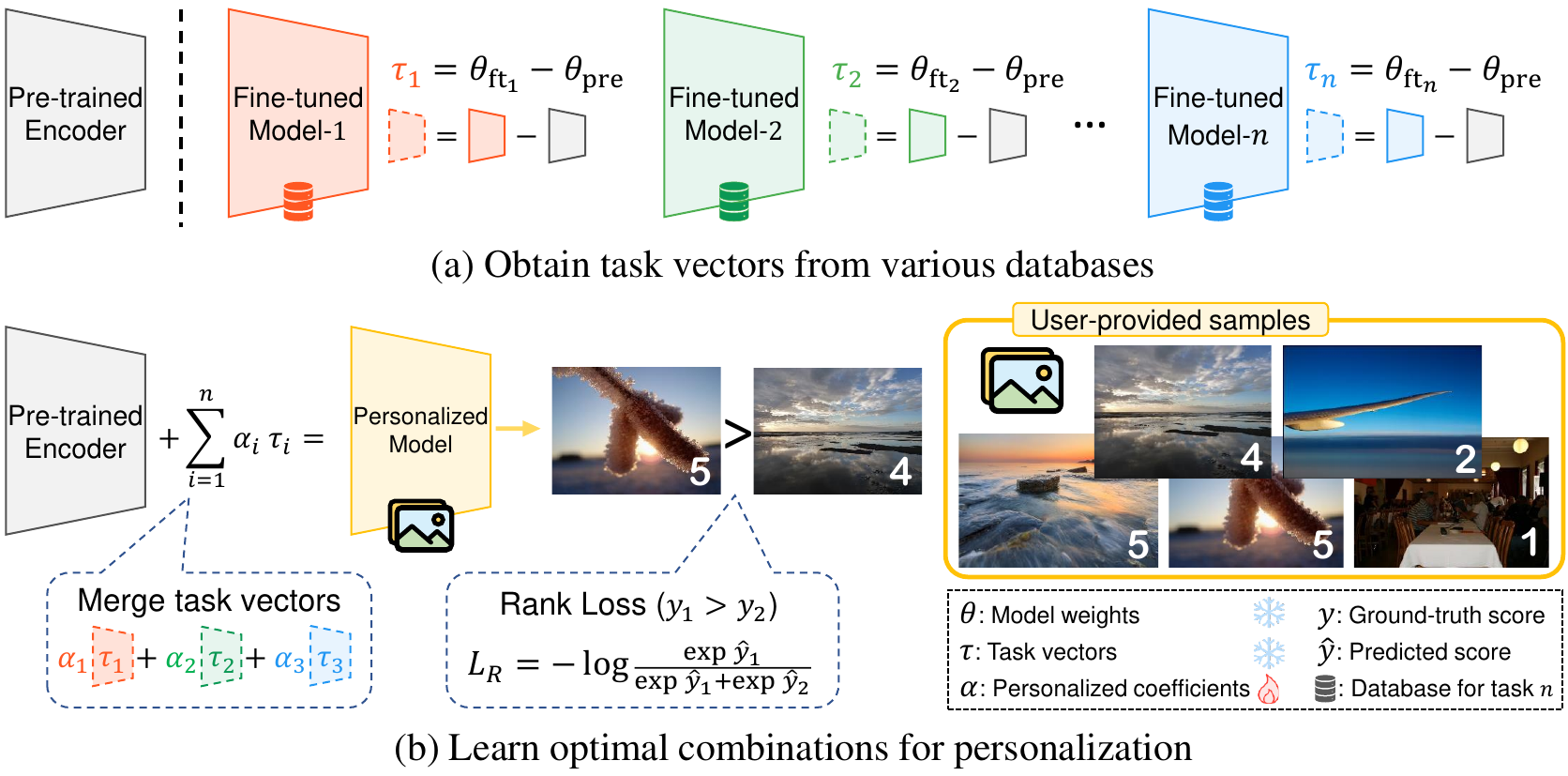}
    \vspace{-0.5cm}
    \caption{(a) Multiple models are fine-tuned from a single pre-trained model on different tasks to obtain task vectors. For simplicity, we omit the discussion of layers. (b) Given a small number of user-provided samples, the coefficients corresponding to each task vector are optimized with the rank loss. }
    \vspace{-0.2cm}
    \label{fig:method-main}
    \vspace{-0.4cm}
\end{figure*}

\section{Method}
\vspace{-0.2cm}

\subsection{Obtaining Layer-wise Task Vectors from Various Databases}
To obtain task vectors, we train multiple models with the same architecture on various tasks (\ie, different databases), as illustrated in \Cref{fig:method-main} (a). 
Specifically, we fine-tune the weights from a pre-trained backbone, adjusting the parameters from $\theta_{\text{pre}}$ to $\theta_{\text{ft}_{\scaleto{i}{4.5pt}}}$ for each task $i$. 
Moreover, while the specific roles of model layers in capturing unique features are inherently complex, their distinct contributions are undeniable~\cite{naseer2021intriguing, chefer2021transformer}. 
Thus, we compute task vectors across \emph{all layers}, enhancing the flexibility of our approach for personalization (See appendix Section A.2). 

Accordingly, we expand the concept of original task vectors to include layer-wise differentiation by,
\begin{equation}
    \tau_i^l = \theta_{\text{ft}_{\scaleto{i}{4.5pt}}}^l - \theta_{\text{pre}}^l,
\end{equation}
where $\tau_i^l$ is the task vector for the $l$-th layer of the $i$-th task. 
Each vector comprehensively integrates patterns learned from large databases, acting as basis vectors for robust and precise personalization. 

It is important to note that this process is conducted \emph{only once} and is not required again for personalizing to individual users, thereby not contributing to the count of learnable parameters for personalization.  

\subsection{Learning Optimal Combinations for Personalization}
Given the layer-wise task vectors, our goal is to find a combination of these vectors that accurately reflects a user's aesthetic preferences. 
To achieve this, we introduce learnable coefficients $\alpha_i^l$ for each task vector $\tau_i^l$. 
Once the coefficients are trained, the weights of the personalized model for each layer are derived as,
\begin{equation}
    \theta_{\text{p}}^l = \theta_{\text{pre}}^l + \sum_{i=1}^{n} \alpha_i^l \tau_i^l \ \ \ \text{for } l=1,2,\dots,L,
\end{equation}
where $n$ is the number of tasks, and $L$ is the number of layers in the model. 
We keep all parameters other than the $n\times L$ coefficients \emph{frozen}. 
This is essential since the task vectors contain comprehensive insights into aesthetic preferences, enabling the model to make \emph{well-informed updates}. 
By harnessing the rich information embedded in these vectors, our method efficiently learns complex aesthetic preferences from merely a few user-provided samples.

Our objective function is designed to penalize the misordering of aesthetic scores. 
Specifically, we adopt the Bradley-Terry model~\cite{bradley}, which estimates the probability of pairwise comparison between two samples. 
We frame the prediction of the two scores as a binary classification problem, where the objective is to predict which sample has the higher score. 
The coefficients of each task vector are trained to minimize the following loss function:
\begin{equation}
    L_{\text{R}} = - \log \frac{\exp\hat{y_1}}{\exp\hat{y}_1 + \exp\hat{y}_2} \ \ \ (y_1 \geq y_2), 
\end{equation}
where $\hat{y}_1$ and $\hat{y}_2$ represent the aesthetic score predictions for images $x_1$ and $x_2$, whose ground-truth scores are $y_1$ and $y_2$. 
We prefer this loss function over the Earth Mover Distance (EMD) or the MSE loss, due to its sample efficiency and fast convergence (See appendix Section A.3). 

\subsection{Adaptive Coefficient Initialization}
We observe that appropriately initializing the coefficients plays a crucial role in determining the coefficients that best align with the user's aesthetic preferences.
When a high coefficient is assigned to a specific task vector, it indicates a strong correlation between the user's aesthetic preference and that particular task.
Thus, we adaptively initialize the coefficients based on the zero-shot personalization performances of each image score regression model associated with the task vector. 

To do this, we calculate the Spearman's rank-order correlation coefficient (SROCC)~\cite{srocc} between the user-provided samples and the scores predicted by each image score regression model.
We then apply the softmax function to initialize the coefficients $\alpha$ as,
\begin{equation}
    \alpha_i^l = \frac{e^{\text{SROCC}(S, \theta_{\text{\scriptsize ft}_{\scaleto{i}{4.5pt}}})}}{\sum_{j=1}^n e^{\text{SROCC}(S, \theta_{\text{\scriptsize ft}_{\scaleto{j}{4.5pt}}})}} \ \ \ \text{for } l=1,2,\dots,L,
\end{equation}
where $\text{SROCC}(S, \theta_{\text{ft}_i})$ indicates the SROCC calculated on the user-provided set $S$ using the model with parameters $\theta_{\text{ft}_i}$. 
Note that we initialize $\alpha$ values for a single task $i$ with the same value for all layers.

\vspace{-0.2cm}
\section{Experiments}

\vspace{-0.1cm}
\subsection{Database Overview}
\vspace{-0.2cm}
The key advantage of our approach is the flexibility to utilize multiple image score regression databases freely, even those without annotator tags. 
In our experiments, we fine-tune models using six different databases in \Cref{tab:exp-database-desc}, reserving the AADB~\cite{aadb} and REAL-CUR~\cite{flickeraes} database exclusively for testing. 
While the Flickr-AES~\cite{flickeraes} and PARA~\cite{para} databases do contain annotator tags, our approach does not require distinguishing between individuals. 
Therefore, we treat these two databases as GIAA databases. 
In total, we effectively leverage \textbf{415,043} samples from six databases, marking the most extensive utilization of samples in PIAA within our knowledge. 
Detailed descriptions and performance on these databases are provided in the appendix (See Section A.5). 
However, our approach is not limited to training solely on these six databases, having the scalability to include additional IQA and GIAA databases~\cite{he2023thinking, dpc, paq2piq, ghadiyaram2015massive, ciancio2010no}. 

For testing, we use the following three most widely-used PIAA test sets. 

\noindent \textbf{Flickr-AES}~\cite{flickeraes} database consists of 40,000 images collected from 210 users. 
For testing purposes, we utilized a test set containing 4,737 images rated by 37 users, with each user contributing image collections ranging from 105 to 171.

\noindent \textbf{AADB}~\cite{aadb} database consists of 9,958 images from 190 users. 
We selected a test set from 22 users, each of which contributed individual image collections containing 110 to 190 images.

\noindent \textbf{REAL-CUR}~\cite{flickeraes} database consists of personal image collections from 14 users, totaling 2,871 images.
Each user's collection ranges from 197 to 222 images.

\begin{table}[t]
\centering
\caption{Description of the databases included in our experiments. The AADB and REAL-CUR databases are exclusively reserved as a test database. }
\vspace{-0.3cm}
\begin{tabular}{@{}x{3cm}x{3cm}x{2.5cm}x{3cm}@{}}
\toprule
Database & Regression task & Annotator tag & Number of images \\ \midrule
KonIQ-10K~\cite{koniq10k} & IQA & \xmark & 10,073  \\
SPAQ~\cite{spaq} & IQA & \xmark & 11,125  \\ \midrule
AVA~\cite{ava} & GIAA & \xmark & 255,500 \\
TAD66K~\cite{tad66k} & GIAA & \xmark & 67,125  \\ \midrule
Flickr-AES~\cite{flickeraes}& GIAA/PIAA & \cmark & 40,000  \\
PARA~\cite{para}& GIAA/PIAA & \cmark & 31,220  \\
AADB~\cite{aadb}& GIAA/PIAA & \cmark & 9,958   \\
REAL-CUR~\cite{flickeraes} & PIAA & \cmark & 2,871   \\ \bottomrule
\end{tabular}
\vspace{-0.2cm}
\label{tab:exp-database-desc}
\end{table}


\vspace{-0.3cm}
\subsection{Implementation Details}
To obtain task vectors from the models fine-tuned on each of the six databases, it is essential to start from an identical pre-trained model. 
Recent approaches in IQA~\cite{musiq}, GIAA~\cite{xu2023clip}, and PIAA~\cite{impiaa} emphasize the importance of pre-training on large databases for training powerful image score regression models. 
Thus, we choose the publicly available OpenCLIP models ViT-B/16 and ViT-L/14~\cite{openclip} as the pre-trained model. 
Full details regarding the architectural design and training hyperparameters are provided in the appendix (See Section A.5 and A.6). 
We will also release the codes upon acceptance. 

If not stated otherwise, we use $n=6$ fine-tuned models to derive the task vectors. 
Each model comprises $L=296$ layers when using the ViT-L/14 model and $L=152$ layers when using the ViT-B/16 model. 
While the number of `layers' usually refers to the number of transformer blocks, we include all computational elements within these blocks, such as linear layers and normalization layers. 
The total number of learnable parameters, $n\times L = 1776$ and $n\times L = 912$, is significantly lower compared to a full fine-tuning approach, which would require training over 23 million parameters even for small ResNet-50 models~\cite{resnet}. 
We \emph{do not} consider the layer-wise task vector itself as a learnable parameter, as obtaining these vectors is not repeated for each individual.

\vspace{-0.3cm}
\subsection{Evaluation Metric for PIAA}
\vspace{-0.1cm}
PIAA approaches are often evaluated through the Spearman's rank-order correlation coefficient~\cite{srocc} (SROCC), which is the correlation coefficient between the predicted and ground-truth rank variables. 
Specifically, given the ground-truth and predicted scores $y$ and $\hat{y}$, the SROCC is calculated as,
\begin{equation}
    \text{SROCC} = 1- \frac{6 \sum_{i=1}^{N}(r_i - \hat{r}_i)^2}{N (N^2-1)},
\end{equation}
where $r_i$ and $\hat{r}_i$ represent the rank of the $i$-th sample within the ground-truth and predicted scores, respectively, and $N$ denotes the number of samples. 

\begin{table}[t]
\centering
\caption{Cross-database evaluation on the REAL-CUR~\cite{flickeraes} database. For the baseline approaches, we report the best cross-domain performances.}
\vspace{-0.3cm}
\begin{tabular}{@{}x{3.5cm}x{3cm}x{3cm}@{}}
\toprule
\multirow{2}{*}{Method} & \multicolumn{2}{c}{SROCC} \\ \cmidrule(l){2-3} 
                        & 10-shot    & 100-shot    \\ \midrule
PA-IAA~\cite{paiaa}     &0.443{\small$\pm$0.004}&0.562{\small$\pm$0.013}\\
BLG-PIAA~\cite{blgpiaa} &0.448{\small$\pm$0.007}&0.578{\small$\pm$0.015}\\  
PIAA-SOA~\cite{piaasoa} &0.487{\small$\pm$0.003}&0.589{\small$\pm$0.015}\\ 
TAPP-PIAA~\cite{tapppiaa}&    -                 &0.580\\ 
MTCL~\cite{mtcl}        &0.495{\small$\pm$0.007}&0.599{\small$\pm$0.012}\\ \midrule
Ours                    &\textbf{0.577{\small$\pm$0.005}}&\textbf{0.621{\small$\pm$0.007}}\\ \bottomrule
\end{tabular}
\vspace{-0.3cm}
\label{tab:exp-realcur}
\end{table}

\subsection{Cross-database Evaluation}
PIAA evaluation can be categorized based on whether training and testing occur within the same database (\ie, intra-database evaluation) or involve different databases (\ie, cross-database evaluation). 
In most real-world scenarios involving personal image collections, user-provided samples often diverge from the statistics or tendencies of the training database. 
Therefore, cross-database evaluation becomes pivotal in assessing the generalization capabilities of personalization approaches, which is an aspect that remains relatively \emph{underdeveloped} in comparison to intra-database evaluations. 

For cross-database evaluation, we employ the AADB~\cite{aadb} and REAL-CUR~\cite{flickeraes} databases as the test databases, neither of which were among the six databases from which we derived our task vectors. 
Following established protocols~\cite{flickeraes}, we randomly select $K$ samples for each individual to form the training set, also referred to as the support set, while the remaining images form the test set. 
In our experiments, we set $K$ as either 10-shot or 100-shot. 
Given the potential variability inherent in few-shot learning depending on the chosen $K$ support set, we conduct 10 independent trials for each user and report the average SROCC with the standard deviation. 

\begin{table}[t]
\centering
\caption{Cross-database evaluation on the AADB~\cite{aadb} database. Results marked with~* indicate intra-database performance (\ie, directly trained on the AADB database).}
\vspace{-0.3cm}
\begin{tabular}{@{}x{3.5cm}x{3cm}x{3cm}@{}}
\toprule
\multirow{2}{*}{Method} & \multicolumn{2}{c}{SROCC} \\ \cmidrule(l){2-3} 
                        & 10-shot    & 100-shot    \\ \midrule
PA-IAA~\cite{paiaa}     &0.469{\small$\pm$0.002}&0.524{\small$\pm$0.006}\\
BLG-PIAA~\cite{blgpiaa} &0.486{\small$\pm$0.004}&0.536{\small$\pm$0.006}\\  
PIAA-SOA~\cite{piaasoa} &0.509{\small$\pm$0.003}&0.557{\small$\pm$0.007}\\ 
TAPP-PIAA~\cite{tapppiaa}&    -                 &0.540\\ 
MTCL~\cite{mtcl}        &0.533{\small$\pm$0.004}&0.572{\small$\pm$0.007}\\ \midrule
*TAPP-PIAA~\cite{tapppiaa}&*0.534{\small$\pm$0.004}&*0.612{\small$\pm$0.007}\\ 
*MTCL~\cite{mtcl}       &*0.540{\small$\pm$0.005}&*0.622{\small$\pm$0.007}\\ \midrule
Ours                    &\textbf{0.556{\small$\pm$0.004}}&\textbf{0.654{\small$\pm$0.007}}\\ \bottomrule
\end{tabular}
\vspace{-0.3cm}
\label{tab:exp-aadb}
\end{table}

As illustrated in \Cref{tab:exp-realcur} and \Cref{tab:exp-aadb}, our method significantly surpasses existing PIAA models in cross-database performance for both the 10-shot and 100-shot protocols. 
Remarkably, \emph{it even exceeds the performance of models trained excplicitly on the AADB database}, as compared in \Cref{tab:exp-aadb}. 
Such results underscore the exceptional generalization capabilities of our approach, in contrast to previous methods that are constrained by their inability to leverage GIAA and IQA databases. 
This superior performance is primarily due to the scalability of our approach, which effectively harnesses an extensive amount of 415,043 samples across multiple databases, each contributing unique insights and tendencies.

\begin{table}[t]
\centering
\caption{10-shot and 100-shot personalization results on the Flickr-AES database~\cite{flickeraes}, with the average SROCC of 37 users and standard deviation across 10 trials.}
\vspace{-0.3cm}
\begin{tabular}{@{}x{4cm}x{3cm}x{3cm}@{}}
\toprule
\multirow{2}{*}{Method} & \multicolumn{2}{c}{SROCC} \\ \cmidrule(l){2-3} 
                        & 10-shot    & 100-shot    \\ \midrule
PAM (attribute only)~\cite{flickeraes}          &0.511{\small$\pm$0.004}&0.516{\small$\pm$0.003}\\
PAM (content only)~\cite{flickeraes}            &0.512{\small$\pm$0.002}&0.516{\small$\pm$0.010}\\
PAM~\cite{flickeraes}                           &0.513{\small$\pm$0.003}&0.524{\small$\pm$0.007}\\
PASS~\cite{pass}                                &0.516{\small$\pm$0.003}&0.521{\small$\pm$0.007}\\
USAR-PPR~\cite{usar}                            &0.521{\small$\pm$0.002}&0.544{\small$\pm$0.007}\\
USAR-PAD~\cite{usar}                            &0.520{\small$\pm$0.003}&0.537{\small$\pm$0.003}\\
USAR-PPR\&PAD~\cite{usar}                       &0.525{\small$\pm$0.004}&0.552{\small$\pm$0.015}\\
PA-IAA~\cite{paiaa}                             &0.543{\small$\pm$0.003}&0.639{\small$\pm$0.011}\\
BLG-PIAA~\cite{blgpiaa}                         &0.561{\small$\pm$0.005}&0.669{\small$\pm$0.013}\\
UG-PIAA~\cite{ugpiaa}                           &0.559{\small$\pm$0.002}&0.660{\small$\pm$0.013}\\
PIAA-SOA~\cite{piaasoa}                         &0.618{\small$\pm$0.006}&0.691{\small$\pm$0.015}\\
TAPP-PIAA~\cite{tapppiaa}                       &0.591{\small$\pm$0.007}&0.685{\small$\pm$0.012}\\
IM-PIAA~\cite{impiaa}                           &0.620                  &0.708                  \\
MTCL~\cite{tapppiaa}                            &0.667{\small$\pm$0.005}&0.737{\small$\pm$0.014}\\ \midrule
Ours                                            &\textbf{0.668{\small$\pm$0.004}}&\textbf{0.748{\small$\pm$0.012}}\\ \bottomrule
\end{tabular}
\vspace{-0.3cm}
\label{tab:exp-flickr}
\end{table}

\vspace{-0.2cm}
\subsection{Intra-database Evaluation}
For intra-database evaluation, we test our approach for the 37 individuals in the Flickr-AES~\cite{flickeraes} database, which is the most widely used PIAA database. 
Since the Flickr-AES~\cite{flickeraes} database is one of the six databases used to derive the task vectors, this experiment falls under the category of intra-database evaluation.
However, we do not use the annotator information in the Flickr-AES.

In \Cref{tab:exp-flickr}, we report the average SROCC for each approach along with the standard deviation across 10 independent trials. 
These experiments underscore the effectiveness of training personalized task vectors through the use of multiple fine-tuned models. 
Our approach consistently outperforms existing methods that rely on meta-learning techniques or incorporate additional user attributes (\eg, personality traits) to enhance personalized score predictions.


\section{Analysis and Ablation Studies}
We perform an in-depth analysis and ablation studies on each component of our approach, including model size, scalability, and coefficient initialization. 

\begin{figure}[t]
    \centering
    \includegraphics[width=0.75\linewidth]{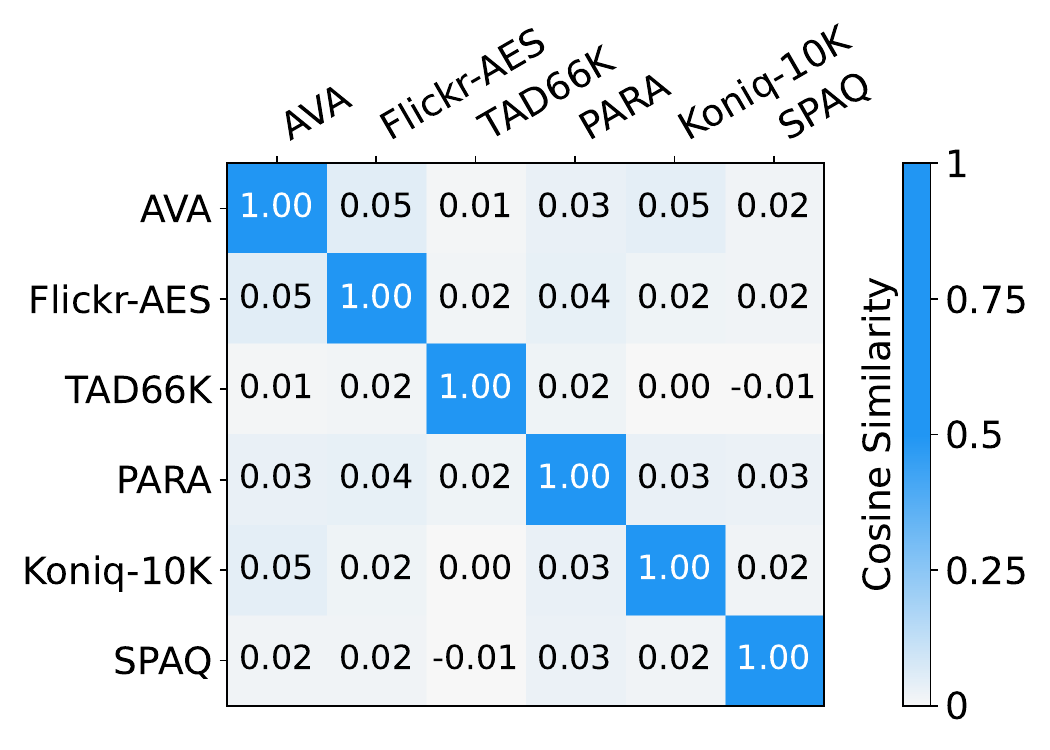}
    \vspace{-0.5cm}
    \caption{Similarity matrix between task vectors derived from each database. The similarities are measured with the cosine similarity. }
    \vspace{-0.2cm}
    \label{fig:exp-cossim}
\end{figure}

\vspace{-0.2cm}
\subsection{Task Similarities Across Databases}
To gain insights into the derived task vectors, we visualize the similarity matrix of the task vectors in \Cref{fig:exp-cossim}. 
For simplicity, we aggregate the $L$ task vectors derived from each layer of a single fine-tuned model as a single task vector by concatenating them into one vector. 
In line with the findings provided by Ilharco~\etal~\cite{taskvector}, task vectors are typically orthogonal, even though the databases are collected with similar intentions (\ie, GIAA and IQA).
This observation highlights that treating each database as a unique task remains a valid choice, owing to variations in their data collection methodologies and the diverse nature of the populations from which they are sourced.

\begin{table}[t]
\centering
\caption{Performance  and inference speed of the ViT-B/16 model and the Best-fit FT method (\ie, directly fine-tuning ViTs) on the REAL-CUR~\cite{flickeraes} database. }
\vspace{-0.3cm}
\begin{tabular}{@{}x{3.5cm}x{2.6cm}x{2.6cm}x{1.7cm}@{}}
\toprule
\multirow{2}{*}{Method} & \multicolumn{2}{c}{SROCC} & \multirow{2}{*}{FPS} \\ \cmidrule(l){2-3} 
                        & 10-shot    & 100-shot    \\ \midrule
Best-fit FT             &0.303{\small$\pm$0.008}&0.394{\small$\pm$0.008}& 94.41 \\ 
PIAA-SOA~\cite{piaasoa} &0.487{\small$\pm$0.003}&0.589{\small$\pm$0.015}& - \\ 
MTCL~\cite{mtcl}        &0.495{\small$\pm$0.007}&0.599{\small$\pm$0.012}& 359.05\\ \midrule
Ours (ViT-B/16)         &0.562{\small$\pm$0.004}&0.607{\small$\pm$0.007}& \textbf{417.95}\\
Ours                    &\textbf{0.577{\small$\pm$0.005}}&\textbf{0.621{\small$\pm$0.007}}& 94.41\\ \bottomrule
\end{tabular}
\vspace{-0.4cm}
\label{tab:exp-vitb}
\end{table}

\subsection{Architecture Choices and Variations}
\noindent\textbf{Selecting the architecture.}
Weight mixing techniques, including task vectors~\cite{wiseft, modelsoup, taskvector}, have been recognized for their effectiveness when fine-tuning models pre-trained on large databases, such as the LAION-5B~\cite{laion} or DataComp~\cite{datacomp}.
Vision Transformers (ViT)~\cite{vit} is a suitable architecture for accommodating web-scale databases, in contrast to ResNets~\cite{resnet}, which often struggle to achieve high performance despite training on large databases. 
Therefore, given its scalability and performance advantages, we choose the publicly available pre-trained ViT as our architecture for our experiments. 
Regardless of the chosen architecture, our parameter-efficient approach only trains a minimal set of coefficients corresponding to each layer-wise task vector. 

\vspace{0.1cm}
\noindent\textbf{Scaling down for efficient computation.}
While our primary experiments focus on combining task vectors derived from ViT-L/14 models, our approach is adaptable to smaller models as well. 
As shown in \Cref{tab:exp-vitb}, our approach yields strong results even on smaller models like ViT-B/16, which are well-suited for deployment on commercial computers. 
Although fast inference was not our primary goal, we achieve real-time inference speeds ({\small$\geq$} 30 FPS) on the RTX 3090, as our weight mixing technique combines multiple models into a single ViT model, eliminating the need to pass through multiple models. 
Acceleration of ViTs~\cite{lin2021fq, yuan2022ptq4vit, liu2021post} beyond our implementation is also actively supported in recent frameworks~\cite{deepspeed} and communities~\cite{pytorch}, which ensures the practicality of our approach across a wide range of computing platforms such as mobile devices. 

\vspace{0.1cm}
\noindent\textbf{Vision Transformers without task vectors.}
Our experimental results in \Cref{tab:exp-vitb} demonstrate that the performance gains are not solely attributed to the inherent strengths of the ViT architecture. 
When directly fine-tuning the aesthetic assessment model that best matches the user's preference (\ie, Best-fit FT), we observe distinctly low personalization performances despite using a large pre-trained model. 
This highlights the pivotal role of keeping the layer-wise task vectors frozen during personalization, as altering the vectors would indicate \emph{discarding} the rich preferences learned from a large database.

\begin{figure}[t]
    \centering
    \includegraphics[width=0.65\linewidth]{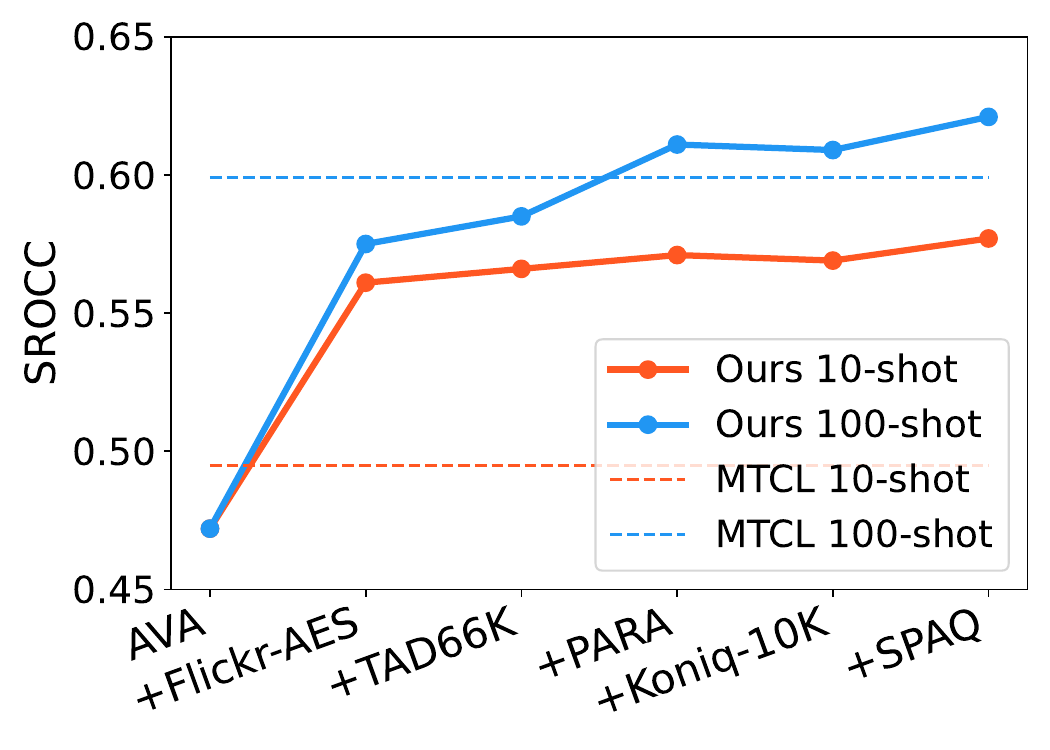}
    \vspace{-0.6cm}
    \caption{Performance gains with respect to the number of databases used. The dotted lines indicate the previous state-of-the-art performance.}
    \vspace{-0.5cm}
    \label{fig:exp-ntasks}
\end{figure}

\vspace{-0.3cm}
\subsection{Scalability of Our Approach}
\noindent\textbf{How does each database contribute? }
We assess the scalability of our approach by varying the number of task vectors.
Specifically, we evaluate our approach on the REAL-CUR database~\cite{flickeraes} by increasing the number of databases used. 
Starting from the GIAA model trained on the AVA~\cite{ava} database, we accumulate the databases in the following order: Flicker-AES~\cite{flickeraes}, TAD66K~\cite{tad66k}, PARA~\cite{para}, KonIQ-10K~\cite{koniq10k}, and SPAQ~\cite{spaq}. 
As demonstrated in \Cref{fig:exp-ntasks}, both 10-shot and 100-shot performance significantly improves as the number of task vectors increases. 
This indicates that our scalable approach benefits from the expansion of task vectors, which allows us to capture a broader range of image characteristics and nuances. 

\vspace{0.1cm}
\noindent\textbf{How can we increase the number of task vectors? }
The advantage of our approach can be amplified with the availability of additional image score regression databases. 
For example, scaling up the number of task vectors can be further pursued by incorporating readily available IQA~\cite{ghadiyaram2015massive, paq2piq, ciancio2010no} and GIAA~\cite{dpc, he2023thinking} databases not included in our current study.
Additionally, partitioning large databases, such as AVA~\cite{ava} and PaQ-2-PiQ~\cite{paq2piq}, can be an alternative method for generating additional task vectors. 

These strategies for obtaining a diverse set of task vectors may lead to performance gains in the 10-shot evaluation, which is considered highly challenging due to the limited number of user-provided samples.  
However, we leave this as future work, given that our current \emph{work already utilizes the largest number of samples for PIAA}, 10 times larger than the most widely used and highly curated Flickr-AES database~\cite{flickeraes}. 

\begin{figure}[t]
    \centering
    \includegraphics[width=0.55\linewidth]{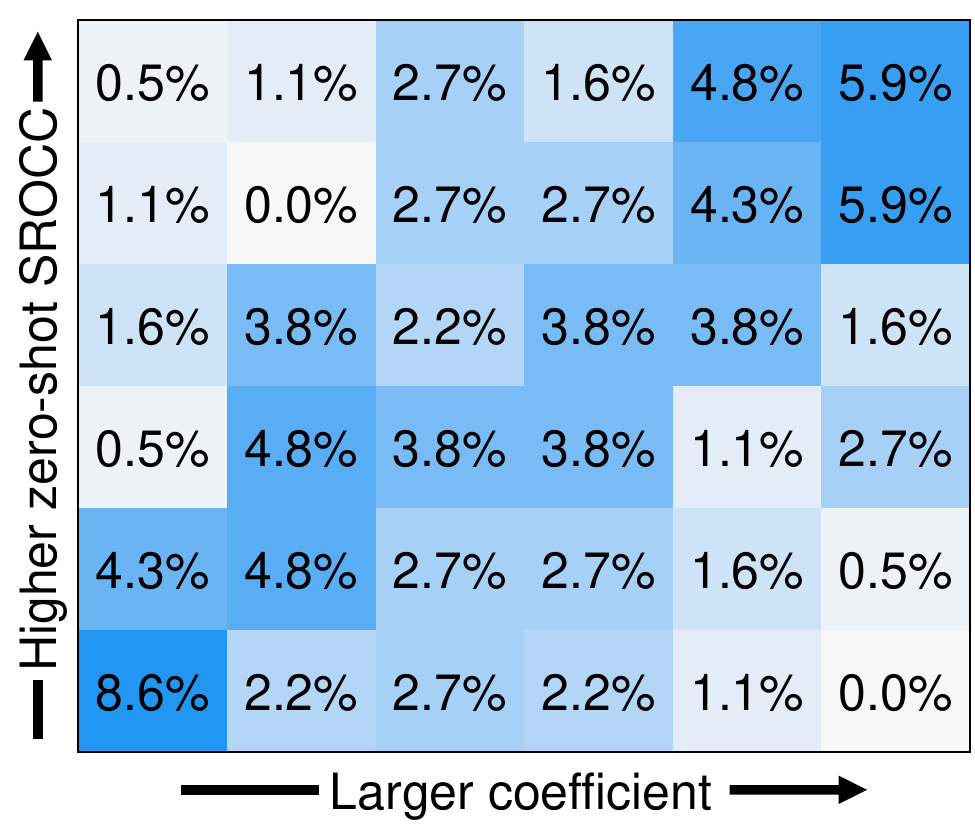}
    \caption{2D-histogram of coefficients and zero-shot SROCC pairs. The size of the coefficients are correlated with the relevance of the task. }
    \vspace{-0.6cm}
    \label{fig:exp-init}
\end{figure}

\vspace{-0.2cm}
\subsection{Task Relevance and Coefficient Initialization}
We conduct an analysis to better understand the distinguishing characteristics of coefficients, specifically, what sets apart those with high values from those with low values.
For a straightforward examination, we calculate the average coefficient for each task (\ie, database), averaged across all the layers of the model, which results in six average coefficients for each user ($n=6$). 
Subsequently, we assess the six zero-shot SROCC scores associated with the average coefficients and plot the histogram of the coefficient-SROCC pairs from all test users in \Cref{fig:exp-init}.
We observe that coefficients for tasks with a high zero-shot SROCC are trained to have high values, while those for less-relevant tasks (\ie, low zero-shot SROCC) exhibit low values. 
This analysis justifies our adaptive initialization of the coefficients, which assigns large values to coefficients associated with closely related tasks. 

We investigate two alternative initialization strategies: a uniform initialization and a best-fit initialization. 
Both cases can be viewed as an extreme case of the adaptive initialization, which can be written in a generalized form:
\begin{equation}
    \alpha_i^l = \frac{e^{\text{SROCC}(S, \theta_{\text{\scriptsize ft}_{\scaleto{i}{4.5pt}}}) / T}}{\sum_{j=1}^n e^{\text{SROCC}(S, \theta_{\text{\scriptsize ft}_{\scaleto{j}{4.5pt}}}) / T}} \ \ \ \text{for } l=1,2,\dots,L,
\end{equation}
where $T$ is the temperature for the softmax. 
The uniform initialization corresponds to the case where the $T\to \infty$, while the best-fit initialization is when $T\to 0$. 
As demonstrated in \Cref{tab:exp-init}, the adaptive initialization outperforms other initialization strategies. 
Nevertheless, we consider further exploration of the initialization strategies as a potential avenue for future research.

\begin{table}[t]
\centering
\caption{Comparison between three different coefficient initialization strategies tested on the REAL-CUR~\cite{flickeraes} database.}
\vspace{-0.3cm}
\begin{tabular}{@{}x{3.5cm}x{3cm}x{3cm}@{}}
\toprule
\multirow{2}{*}{Method} & \multicolumn{2}{c}{SROCC} \\ \cmidrule(l){2-3} 
                       & 10-shot    & 100-shot    \\ \midrule
Uniform init. ($T\to\infty$)    &0.520{\small$\pm$0.002}&0.602{\small$\pm$0.002}\\
Best-fit init. ($T\to0$)        &0.464{\small$\pm$0.012}&0.596{\small$\pm$0.002}\\
Adaptive init. ($T=1$)          &\textbf{0.577{\small$\pm$0.005}}&\textbf{0.621{\small$\pm$0.007}}\\ \bottomrule
\end{tabular}
\vspace{-0.4cm}
\label{tab:exp-init}
\end{table}

\section{Limitation}
\vspace{-0.2cm}
We have observed that when the zero-shot SROCC for all six image regression models are low for an individual (\ie, $\text{SROCC}\approx0$), our approach may struggle to identify an optimal combination of task vectors for that user.
This phenomenon suggests that when there is a limited correlation between the user's preferences and the predictions of these models, the task vectors derived from them may be insufficient for finding an appropriate personalized task vector. 

However, we find this case to be extremely rare (See appendix Figure 2-7.), with only \emph{1 among 73} cases having a low SROCC under 0.2, demonstrating the broad coverage of our approach. 
Also, increasing the number of task vectors will effectively mitigate this issue, leveraging the inherent scalability of our approach, which can accommodate a growing number of task vectors to improve personalization even in challenging scenarios. 


\vspace{-0.2cm}
\section{Conclusion}
\vspace{-0.2cm}
We address the scalability issue in PIAA, a critical yet previously unaddressed aspect necessary for enhancing the generalization performance. 
Unlike previous studies limited to training on PIAA databases curated by individual annotators, our approach enables the flexible use of multiple image score regression databases, thereby providing greater scalability and generaliation capabilities.

We demonstrate that GIAA and IQA databases exhibit distinct personalization potentials, which have motivated us to combine GIAA and IQA models to achieve desired behaviors. 
By introducing learnable parameters for optimal model combinations, we enable the precise personalization of models to individual users' aesthetic preferences. 
Our approach significantly outperforms existing approaches, demonstrating exceptional efficacy in generalizing to unseen domains which is a critical requirement for real-world applications. 
This work opens up new avenues for personalized image aesthetic assessment, offering valuable insights and practical solutions for this challenging task.

\vspace{-0.3cm}
\section*{\small Acknowledgement}
\vspace{-0.3cm}
{\scriptsize This work was supported by the Institute for Information \& communications Technology Promotion(IITP) grant funded by the Korea government(MSIT) (No.RS-2019-II190075 Artificial Intelligence Graduate School Program(KAIST)), the National Research Foundation of Korea (NRF) grant funded by the Korea government (MSIT) (No. NRF-2022R1A2B5B02001913), and the Institute of Information \& communications Technology Promotion (IITP) grant funded by the Korea government (MSIT) (RS-2021-II212068, Artificial Intelligence Innovation Hub).}

\clearpage

\setcounter{section}{0}
\renewcommand*{\thesection}{\Alph{section}}
{\huge\textbf{Appendix}}

\section{\large Individual Personalization Results}
\begin{figure}[ht]
    \centering
    \includegraphics[width=0.5\linewidth]{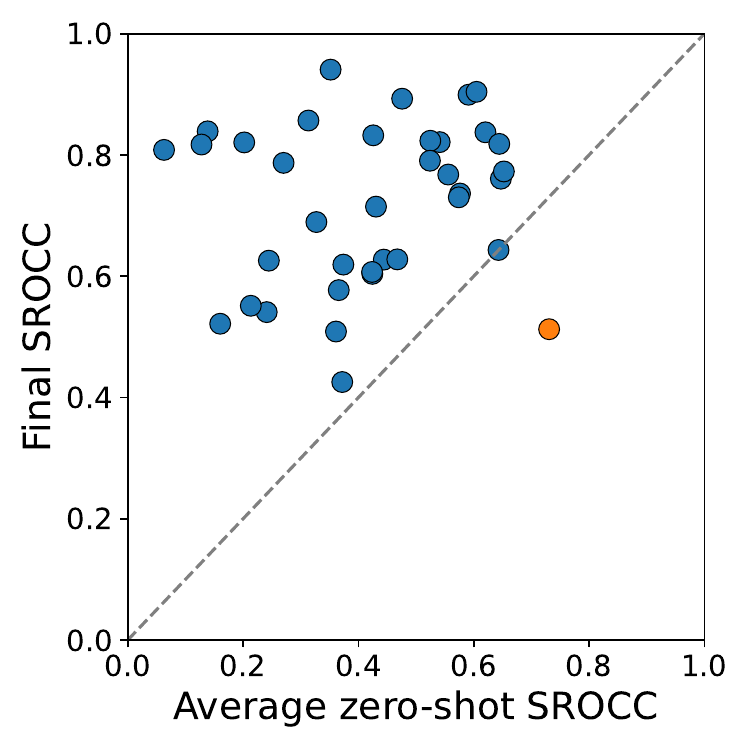}
    \vspace{-0.2cm}
    \caption{Performance gains visualized for each individual in the Flickr-AES database~\cite{flickeraes}. Our approach generally improves the persoanlization performance for individuals. }
    \vspace{-0.3cm}
    \label{fig:supp-zero-to-final}
\end{figure}

\begin{figure*}
    \centering
    \includegraphics[width=\textwidth]{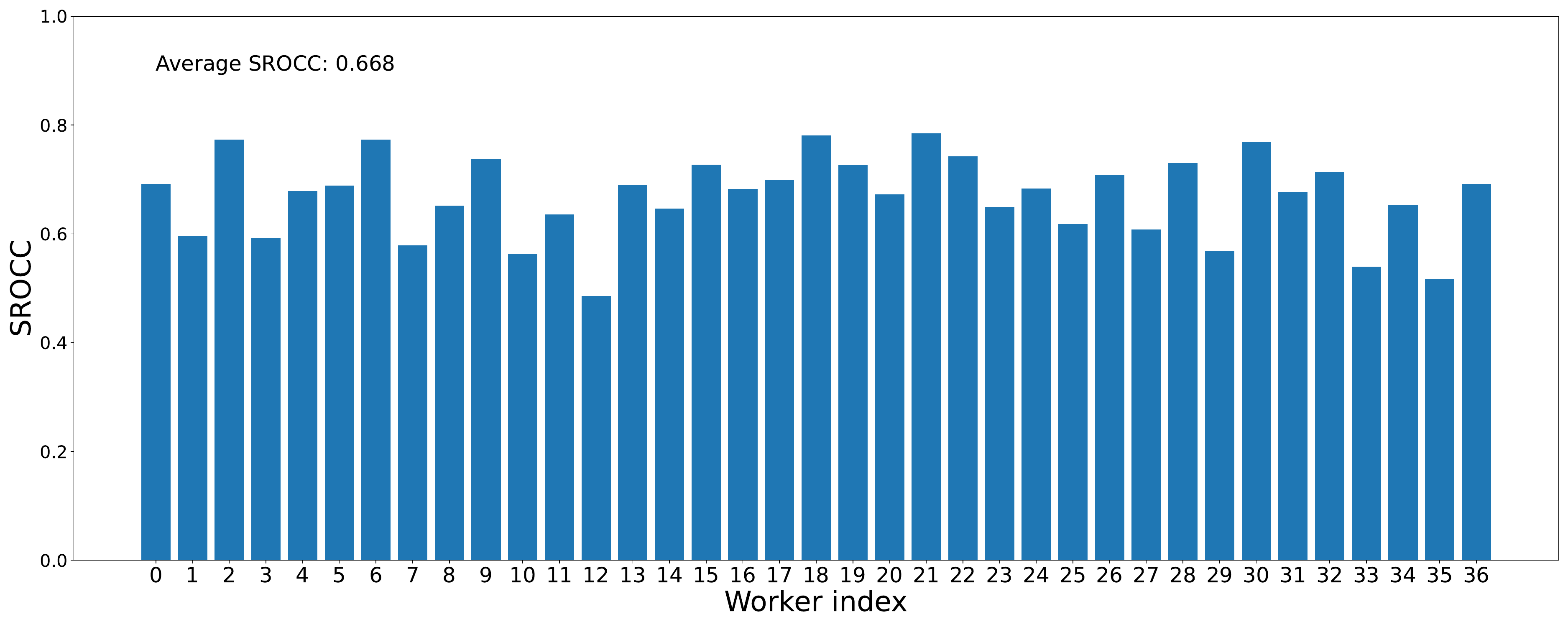}
    \caption{Individual 10-shot personalization performance of the individuals in the Flickr-AES database~\cite{flickeraes}. }
    \label{fig:supp_aes_k10}
\end{figure*}

\begin{figure*}
    \centering
    \includegraphics[width=\textwidth]{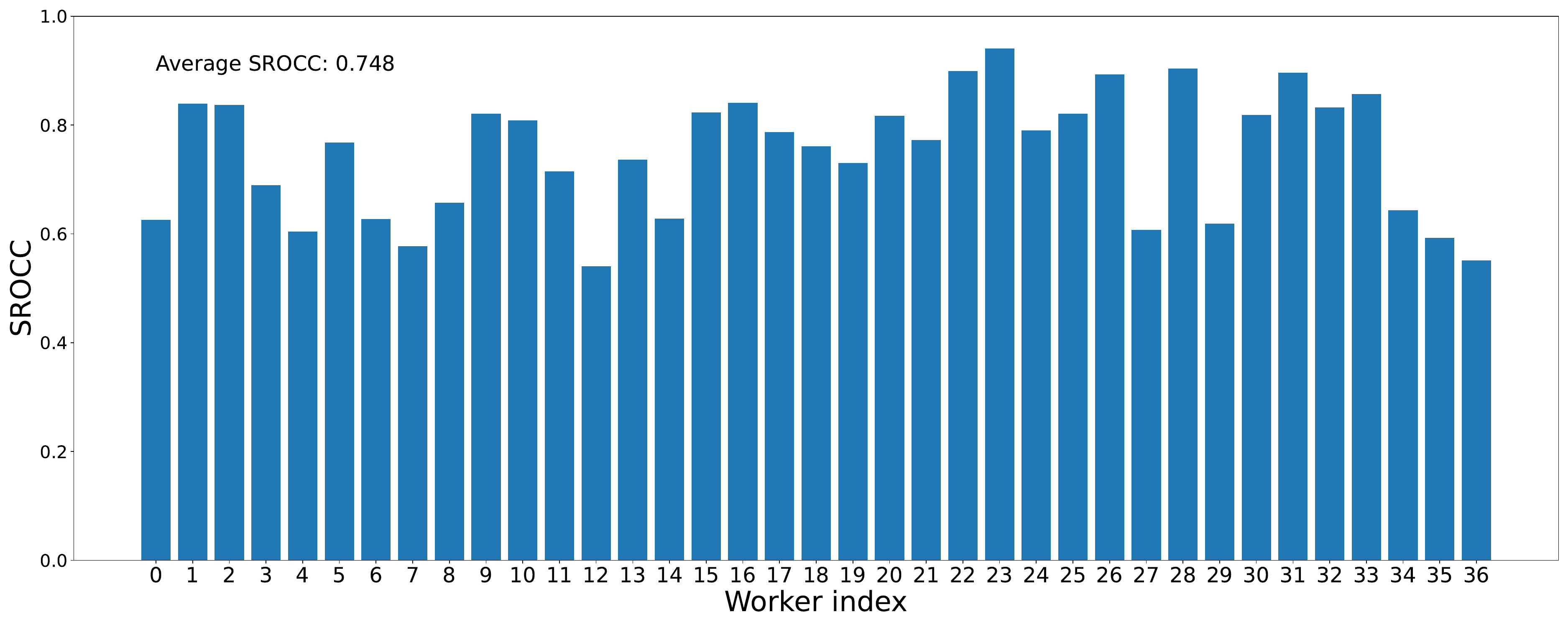}
    \caption{Individual 100-shot personalization performance of the individuals in the Flickr-AES database~\cite{flickeraes}.}
    \label{fig:supp_aes_k100}
\end{figure*}

\begin{figure*}
    \centering
    \includegraphics[width=\textwidth]{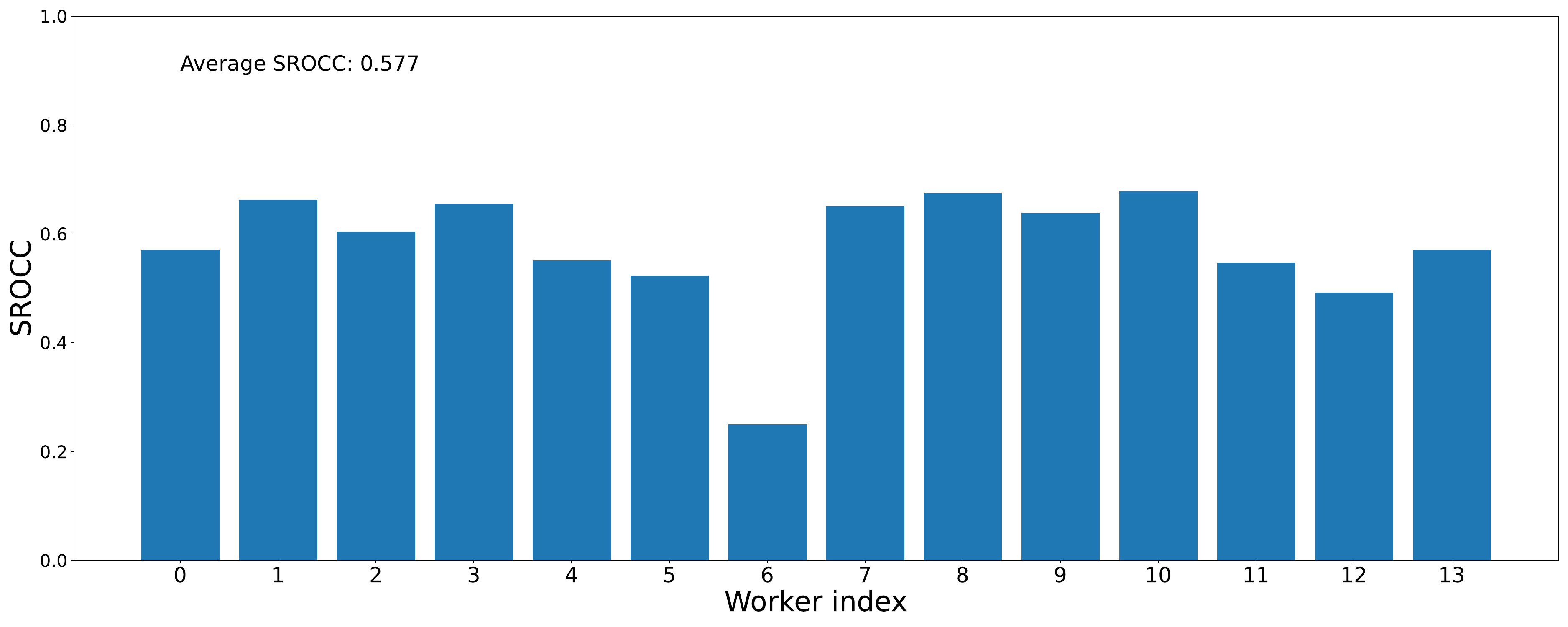}
    \caption{Individual 10-shot personalization performance of the individuals in the REAL-CUR database~\cite{flickeraes}}
    \label{fig:supp_realcur_k10}
\end{figure*}

\begin{figure*}
    \centering
    \includegraphics[width=\textwidth]{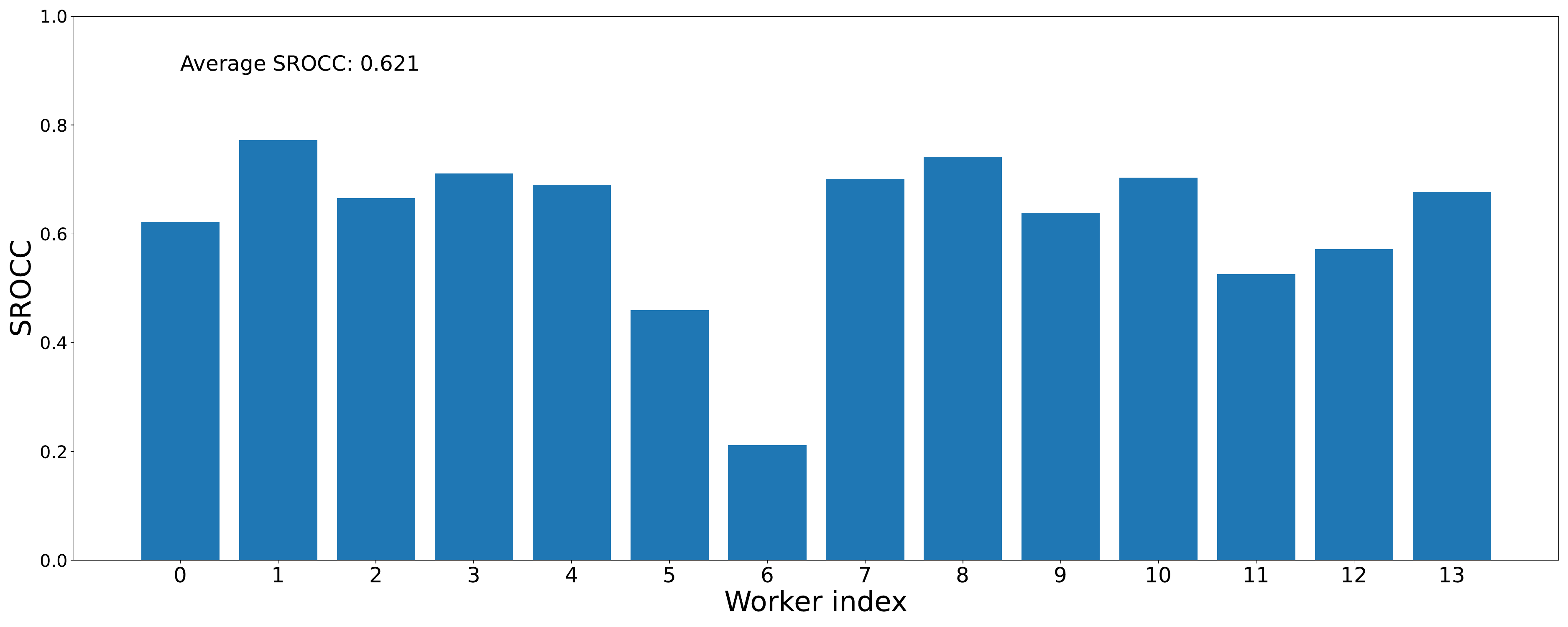}
    \caption{Individual 100-shot personalization performance of the individuals in the REAL-CUR database~\cite{flickeraes}}
    \label{fig:supp_realcur_k100}
\end{figure*}

\begin{figure*}
    \centering
    \includegraphics[width=\textwidth]{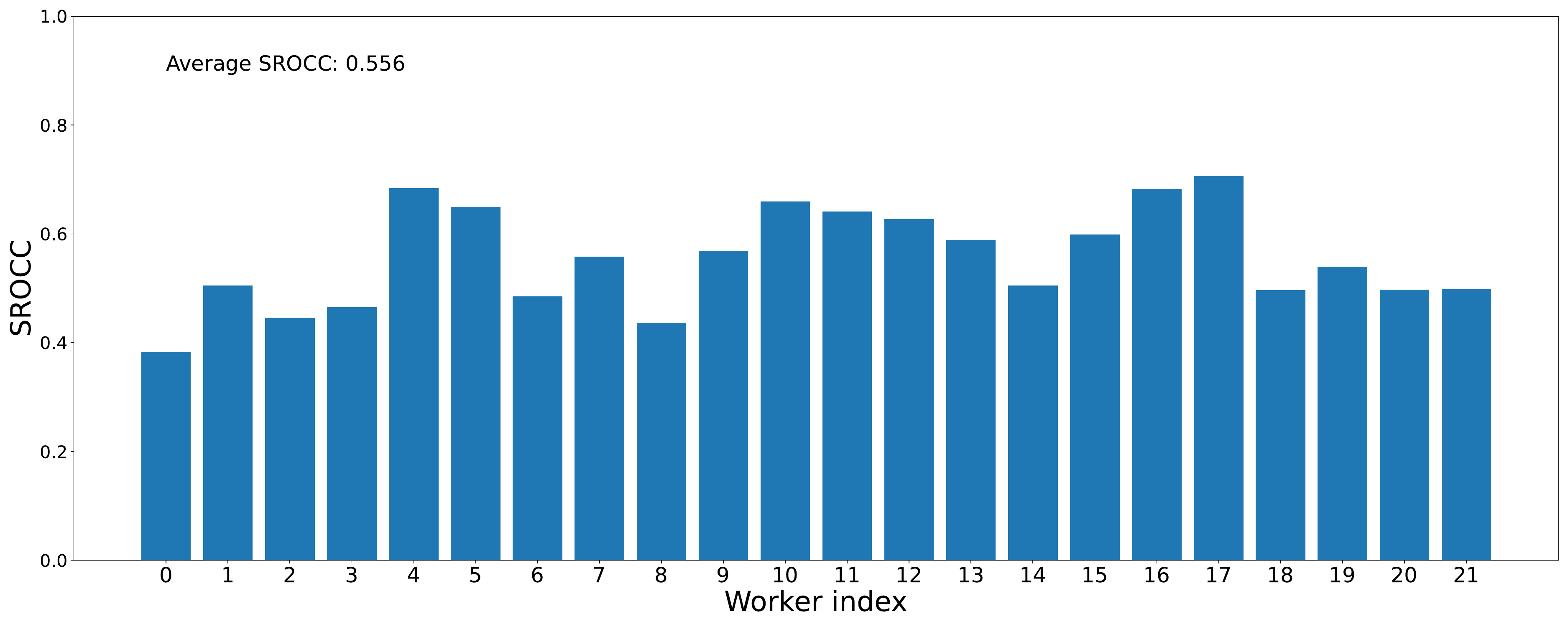}
    \caption{Individual 10-shot personalization performance of the individuals in the AADB database~\cite{aadb}}
    \label{fig:supp_aadb_k10}
\end{figure*}

\begin{figure*}
    \centering
    \includegraphics[width=\textwidth]{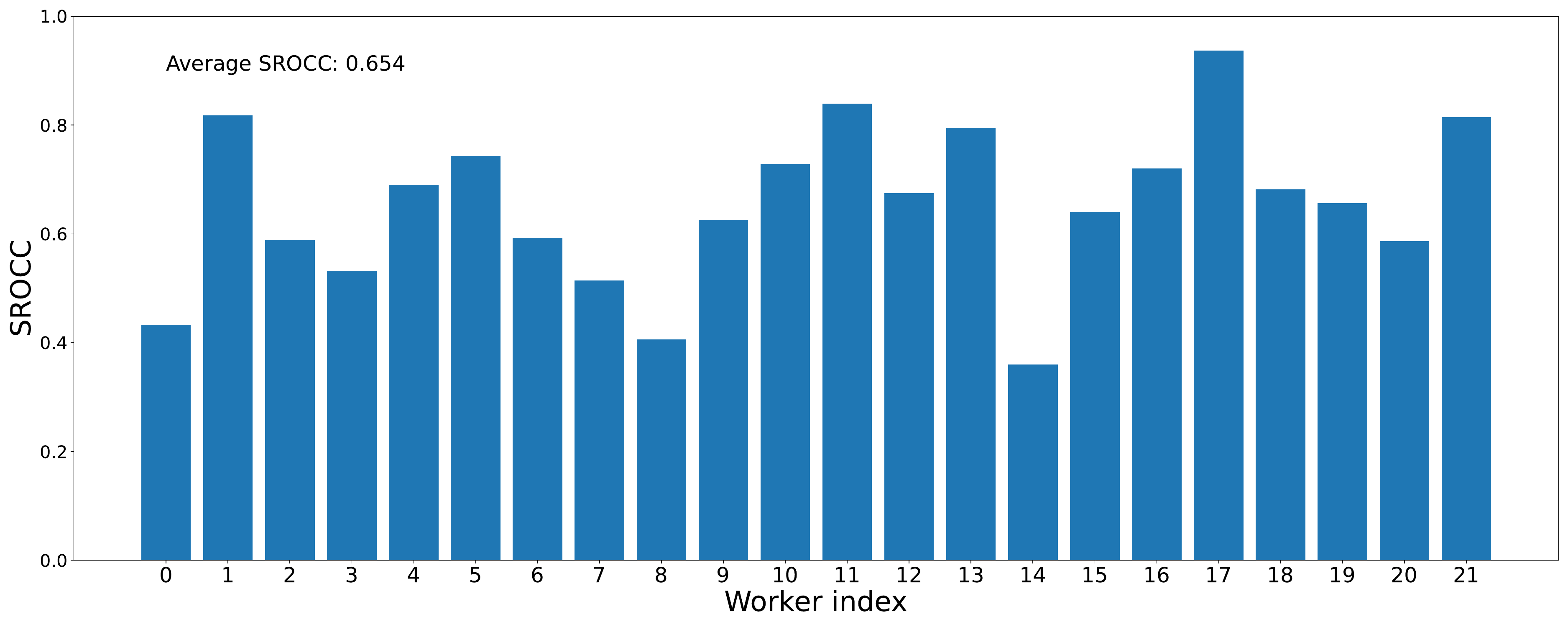}
    \caption{Individual 100-shot personalization performance of the individuals in the AADB database~\cite{aadb}}
    \label{fig:supp_aadb_k100}
\end{figure*}

The personalization of aesthetic assessment models is commonly evaluated based on the average personalization performance across multiple individuals, allowing us to test the approach on various individual preferences. 
While achieving a high average performance is important, it is equally crucial to ensure that personalization is effective for every individual, without exceptions.
To address this, we present a visualization of the performance improvements observed after personalizing the aesthetic model for 37 individuals from the Flickr-AES database~\cite{flickeraes}. 
Specifically, we calculate the average zero-shot Spearman rank-order correlation coefficient (SROCC) across the six models (fine-tuned on six databases) as the baseline performance before personalization. 
We then plot the performance enhancements achieved after personalization with the 100 user-provided samples (\ie, 100-shot).
The average zero-shot SROCC can be understood as a simple logit-ensemble of the six models.
As shown in \Cref{fig:supp-zero-to-final}, it is clear that training the coefficients of the task vectors leads to a significant increase in personalization scores, demonstrating the effectiveness of our approach in adapting to the diverse personal preferences of individuals. 
We also add full details of individual persoanlization performances in \Cref{fig:supp_aes_k10,fig:supp_aes_k100,fig:supp_aadb_k10,fig:supp_aadb_k100,fig:supp_realcur_k10,fig:supp_realcur_k100}.

\section{\large Importance of Layer-wise Task Vectors}
We conduct an additional ablation study to assess the importance of deriving task vectors for each layer, as opposed to the method proposed by Ilharco~\etal~\cite{taskvector}, which derives a single task vector for each task.
When a single task vector is derived for each task, the total number of task vectors is reduced from $n\times L=1776$ to just $n=6$, where $n$ represents the number of tasks, and $L$ denotes the number of layers in the ViT-L/14~\cite{vit} model.
The capacity of this approach, which we call as the layer-agnostic task vectors, is limited due to the extremely small number of trainable parameters.
As seen in \Cref{tab:supp-layer-agnostic}, training the coefficients of the layer-agnostic task vectors results in poor personalization performance, underlining the flexibility and importance of layer-wise task vectors.

\section{\large Comparing Different Loss Functions}
Given the distinct nature of our approach, which does not require training the entire model parameters, we explore various loss functions suitable for training the learnable coefficients.
In the realm of general image aesthetic assessment (GIAA), two prevalent loss functions are the Earth Mover's Distance (EMD) loss and the Mean Squared Error (MSE) loss.
However, EMD loss is exclusively applicable to databases like the AVA database~\cite{ava}, which collect multiple scores per image, making it less suited for personalization tasks involving individual preferences.

Considering an alternative, we turn to the MSE loss, known for its simplicity and effectiveness. 
Additionally, we delve into the Bradley-Terry model~\cite{bradley}, extensively utilized in recent reinforcement learning with human feedback (RLHF) approaches for training reward models.
Our findings indicate that the rank loss derived from the Bradley-Terry model excels in sample efficiency, leveraging a combination of samples for training. 
Additionally, we find that the convergence of the MSE loss is slower than the EMD loss, requiring twice as long training steps to achieve the accuracy reported in \Cref{tab:supp-layer-agnostic}.

\begin{table}[t]
\centering
\caption{Comparison between the performance when deriving task vectors in a layer-agnostic and layer-wise manner.}
\vspace{-0.3cm}
\begin{tabular}{@{}x{4cm}x{3cm}x{3cm}@{}}
\toprule
\multirow{2}{*}{Method} & \multicolumn{2}{c}{SROCC} \\ \cmidrule(l){2-3} 
                        & 10-shot    & 100-shot    \\ \midrule
PA-IAA~\cite{paiaa}     &0.443{\small$\pm$0.004}&0.562{\small$\pm$0.013}\\
BLG-PIAA~\cite{blgpiaa} &0.448{\small$\pm$0.007}&0.578{\small$\pm$0.015}\\  
PIAA-SOA~\cite{piaasoa} &0.487{\small$\pm$0.003}&0.589{\small$\pm$0.015}\\ 
TAPP-PIAA~\cite{tapppiaa}&    -                 &0.580\\ 
MTCL~\cite{mtcl}        &0.495{\small$\pm$0.007}&0.599{\small$\pm$0.012}\\ \midrule
Ours (MSE loss)         & 0.553{\small$\pm$0.002}& 0.600{\small$\pm$0.003}\\
Ours (layer-agnostic)   &0.438{\small$\pm$0.003}&0.543{\small$\pm$0.003}\\ 
Ours                    &\textbf{0.572{\small$\pm$0.005}}&\textbf{0.621{\small$\pm$0.007}}\\ \bottomrule
\end{tabular}
\vspace{-0.5cm}
\label{tab:supp-layer-agnostic}
\end{table}

\section{\large Detailed Architectural Designs}
\label{sec-archi}
Task vector arithmetic, as described in~\cite{taskvector}, is performed by adding or subtracting the fine-tuned weights in an element-wise manner. 
Therefore, when combining multiple task vectors, they must be of the same size. 
However, architectural designs vary across tasks, as specialized strategies are employed for each task. 
Additionally, different databases have varying score ranges; some range from 1 to 10, while others range from 1 to 5.

To address these challenges, we present a model architecture that can accommodate all tasks and facilitate the derivation of task vectors. 
As mentioned in Section 5 of the manuscript, we select Vision Transformers~\cite{vit} (ViT) as the suitable backbone architecture for deriving task vectors. 
Building upon the findings by Hosu~\etal~\cite{hosu}, we add two linear layers with the final output channel size fixed to 10. 
The final regression score is calculated as the dot product between the 10 output scores and a 10-dimensional template vector, which is a list ranging from 1 to 10. 
For tasks with score ranges from 1 to 5, we adjust the 10-dimensional template vector to include values from 0.5 to 5 in 0.5 increments.
In this way, we are able to fix the number of parameters for all databases. 
The summary of these architectural details is as follows:
\begin{lstlisting}[language=Python,]
image_score_regression_model: {
    embed_dims: 768,
    backbone: {
        image_size: 224,
        layers: 24,
        width: 1024,
        patch_size: 14
    },
    head: {
        hidden_dim: 512,
        non_linearaity: GELU,
        dropout: 0.5,
        output_dim: 10,
        output_layer: Sigmoid
    }
}
\end{lstlisting}
\vspace{-0.4cm}

\section{\large Characteristics of the Training Database}
In this section we provide descriptions on the data size, intentions, and collection schemes for each of the six training databases used to derive task vectors.

\noindent\textbf{AVA.} 
The AVA database~\cite{ava} is the largest database for training models for general image aesthetic assessment (GIAA), containing over 200,000 images. 
These images, along with their corresponding aesthetic scores, are collected from a digital photography contest community, \href{www.dpchallenge.com}{DPchallenge}.
Each image contains semantic tags and photographic style tags, along with a aesthetic score \emph{vote} from multiple individuals. 
The number of votes per image ranges from 78 to 549, making the aesthetic score labels a robust representation of the community's average opinion.
In our study, we exclusively utilize the aesthetic scores to train a single model. 
However, leveraging the rich semantic and photographic style tags to segment this vast database into multiple subsets represents a promising avenue for future research.

\noindent\textbf{Flickr-AES.}
Utilizing images collected from the large photo collection \href{www.flickr.com}{Flickr}, the Flickr-AES database~\cite{flickeraes} offers aesthetic score ratings for 40,000 images, with each image rated by 5 AMT annotators. 
Each image in this database has been rated by 5 AMT annotators. 
This diverse collection is suitable for both GIAA and PIAA tasks, as it includes tags for the annotator IDs with each image. 
When the dataset is restructured from the perspective of each individual annotator, it shows that 210 AMT annotators have each labeled between 105 to 171 images. 
However, we treat this database primarily for GIAA, as our scalable method for training PIAA models does not necessitate the use of annotator tags.

\noindent\textbf{TAD66K.}
TADK66K~\cite{tad66k} is one of the recently proposed databases for GIAA, specifically curated to ensure a balanced representation of various themes such as sunset, sea, winter, and street.
Furthermore, different evaluation criteria were given to the annotators for different themes, as the aesthetic elements may differ for different themes.
More than 60,000 images were collected and annotated by 1200 annotators.

\noindent\textbf{PARA.}
The PARA database~\cite{para}, meeting the needs of recent PIAA approaches, has gathered aesthetic scores of more than 30,000 images from 438 annotators, marking the largest involvement of annotators in any such database to date.
The database also includes rich information about the annotators, such as personality traits, photographic and artistic experience, emotion, age and gender. 
Although these detailed attributes offer the potential to categorize annotators, our approach does not utilize these additional attributes for the derivation of task vectors.

\noindent\textbf{KonIQ-10K.}
KonIQ-10K~\cite{koniq10k} stands as one of the pioneering large-scale image quality assessment (IQA) databases, distinct in its approach of collecting a variety of images and annotating their quality, rather than creating low-quality images by distorting high-quality ones. 
The 10,000 images in this database are labeled based on aspects of photographic quality such as brightness, colorfulness, and contrast, which is a criteria that differ from those used in GIAA databases. 
These alternative criteria provide unique task vectors, which are crucial for covering a wide range of personal preferences.

\noindent\textbf{SPAQ.}
Specifically targeting the unique domain of smartphone photography, the SPAQ database~\cite{spaq} comprises a collection of 11,125 images captured using 66 different smartphones. 
Each image in this database is labeled according to its photographic quality. 
These collections offer domain-specific insights, particularly focusing on smartphone photography, thereby broadening the range of domains encompassed through the combination of task vectors.

\section{\large Training Hyper-parameters}
Our method comprises two key phases designed to optimize personalized model training. 
The initial phase focuses on generating task vectors for each database, laying the groundwork for our approach. 
This step, although time-intensive, is performed only once, enabling the model to be easily adapted for various individuals thereafter. 
The second phase is centered on fine-tuning the model for individual-specific requirements. Here, our objective is to minimize training time without compromising on performance. 
By investing in a robust first phase, we ensure the subsequent personalization process is both efficient and scalable, catering to multiple individuals with a single, comprehensive training cycle. 

In the first stage, we initially train a regression head with the backbone parameters fixed, followed by fine-tuning the backbone. 
Note that this stage conducted \emph{only once} and is not required again for personalizing to individual users. 
Furthermore, we will make available the model weights fine-tuned on each database to support future research efforts. 
\begin{lstlisting}[language=Python,]
task_vector_acquisition_phase: {
    optimizer: AdamW,
    lr_schedule: Cosine annealing,
    train_head:{
        start_lr: 1.5e-5,
        end_lr: 1.5e-6,
        batch_size: 128,
        steps: 60,000
    }
    fine_tune_backbone:{
        start_lr: 1.5e-6,
        end_lr: 1.5e-7,
        batch_size: 32,
        steps: 5,000
    }
}
\end{lstlisting}

During the personalization phase, we keep all freeze parameters frozen, including the task vectors and pre-trained weights, and concentrate solely on training the personalization coefficients. 
\begin{lstlisting}[language=Python,]
personalization_phase: {
    start_lr: 1.0e-2,
    end_lr: 1.0e-3,
    lr_schedule: Cosine annealing,
    batch_size: 32,
    steps: 500
}
\end{lstlisting}
Note that the presonalization phase only takes \textbf{500 iterations} due to the small number of trainable parameters.

\begin{table}[ht]
\centering
\begin{minipage}[t]{0.485\linewidth}
\caption{GIAA performance on the AVA database~\cite{ava}}
\vspace{-0.3cm}
\begin{tabular}{@{}x{3.2cm}x{1.2cm}x{1.2cm}@{}}
\toprule
Method               & PLCC  & SROCC \\ \midrule
NIMA~\cite{nima}     & 0.636 & 0.612 \\
Hosu~\etal~\cite{hosu}&0.757 & 0.756 \\
PA-IAA~\cite{paiaa}  &   -   & 0.666 \\
HLA-GCN~\cite{hlagcn}& 0.687 & 0.665 \\
MUSIQ-single~\cite{musiq}& 0.731 & 0.719 \\
MUSIQ~\cite{musiq}   & 0.738 & 0.726 \\
VILA-R~\cite{vila}   & 0.774 & 0.774 \\ 
TANet~\cite{tad66k}  & 0.765 & 0.758 \\
TAVAR~\cite{tavar}   & 0.736 & 0.725 \\
CSKD~\cite{xu2023clip}&0.779 & 0.770 \\ \midrule
Ours (ViT-B/16)      & 0.780 & 0.781 \\ 
Ours (ViT-L/14)      & \textbf{0.804} & \textbf{0.808} \\ \bottomrule
\end{tabular}
\label{tab:ava}
\end{minipage}
\hspace{0.1cm}
\begin{minipage}[t]{0.485\linewidth}
\caption{GIAA performance on the TAD66K database~\cite{tad66k}}
\vspace{-0.3cm}
\begin{tabular}{@{}x{3.2cm}x{1.2cm}x{1.2cm}@{}}
\toprule
Method               & PLCC  & SROCC \\ \midrule
PAM~\cite{flickeraes}&0.440 & 0.422 \\
NIMA~\cite{nima}     &0.405 & 0.390 \\
HLA-GCN~\cite{hlagcn}&0.493 & 0.486 \\ 
TANet~\cite{tad66k}  &\textbf{0.531} & \textbf{0.513} \\ \midrule
Ours (ViT-B/16)      &0.521 & 0.491 \\ 
Ours (ViT-L/14)      &0.523 & 0.492 \\ \bottomrule
\label{tab:tad66k}
\end{tabular}
\vspace{-0.25cm}
\caption{GIAA performance on the PARA database~\cite{para}}
\vspace{-0.3cm}
\begin{tabular}{@{}x{3.2cm}x{1.2cm}x{1.2cm}@{}}
\toprule
Method                & PLCC  & SROCC \\ \midrule
PARA~\cite{para}      &0.936 & 0.902 \\
CSKD~\cite{xu2023clip}&\textbf{0.951} & \textbf{0.926} \\ \midrule
Ours (ViT-B/16)       & 0.945 & 0.921 \\ 
Ours (ViT-L/14)       & 0.945 & 0.917 \\ \bottomrule
\end{tabular}
\label{tab:para}
\end{minipage}
\vspace{-0.7cm}
\end{table}
\begin{table}[t]
\centering
\begin{minipage}[h]{0.485\linewidth}
\vspace{-0.2cm}
\caption{IQA performance on the KonIQ-10K database~\cite{koniq10k}}
\vspace{-0.3cm}
\begin{tabular}{@{}x{3cm}x{1.2cm}x{1.2cm}@{}}
\toprule
Method               & PLCC  & SROCC \\ \midrule
BRISQUE~\cite{brisque}&0.681 & 0.665 \\
ILNIQE~\cite{ilniqe} &0.523 & 0.507 \\
SFA~\cite{sfa}       & 0.872 & 0.856 \\
DBCNN~\cite{dbcnn}   & 0.884 & 0.875 \\
MetaIQA~\cite{metaiqa}& 0.887 & 0.805 \\
BIQA~\cite{biqa}     & 0.917 & 0.906 \\ 
MUSIQ~\cite{musiq}   & 0.928 & 0.916 \\ 
CLIP-IQA~\cite{clipiqa}& 0.909 & 0.895 \\ \midrule
Ours (ViT-B/16)      & \textbf{0.941} & \textbf{0.925} \\ 
Ours (ViT-L/14)      & 0.933 & 0.918 \\ \bottomrule
\end{tabular}
\label{tab:koniq}
\end{minipage}
\hspace{0.1cm}
\begin{minipage}[h]{0.485\linewidth}
\centering
\vspace{0.2cm}
\caption{IQA performance on the SPAQ database~\cite{spaq}}
\vspace{-0.3cm}
\begin{tabular}{@{}x{3cm}x{1.2cm}x{1.2cm}@{}}
\toprule
Method               & PLCC  & SROCC \\ \midrule
BRISQUE~\cite{brisque}&0.817 & 0.809 \\
ILNIQE~\cite{ilniqe} &0.721 & 0.713 \\
DBCNN~\cite{dbcnn}   & 0.915 & 0.911 \\
Fang~\etal~\cite{spaq}&0.909 & 0.908 \\
BIQA~\cite{biqa}     & \textbf{0.928} & \textbf{0.925} \\
MUSIQ~\cite{musiq}   & 0.921 & 0.917 \\ 
CLIP-IQA~\cite{clipiqa}& 0.866 & 0.864 \\ \midrule
Ours (ViT-B/16)      & 0.882 & 0.914 \\ 
Ours (ViT-L/14)      & 0.874 & 0.912 \\ \bottomrule
\end{tabular}
\label{tab:spaq}
\end{minipage}
\vspace{-0.3cm}
\end{table}

\section{\large Performance on Each GIAA/IQA Database}
In this section, we report the performance of the models trained for each image score regression task. 
While our unified architecture is not explicitly designed for specific tasks, such as image quality assessment (IQA) or general image aesthetic assessment (GIAA), the performance achieved for each task is comparable to existing approaches as demonstrated in \Cref{tab:ava,tab:tad66k,tab:para,tab:koniq,tab:spaq}. 
This demonstrates the effectiveness of the unified architecture described in \Cref{sec-archi}. 
A high performance on each database indicates that the model has successfully learned the characteristics of the databases, including the distribution of scores and preference for certain types of images.
Thus, we can assure that deriving task vectors from these models produces task vectors that accurately represents the characteristics of each database.

Furthermore, we observe that larger models do not always guarantee higher performance, aligning with some of the results reported in recent IQA studies~\cite{clipiqa, musiq}. 
This underscores the importance of effectively harnessing the capabilities of pre-trained models, which is also the central idea of our approach in training the coefficients of task vectors. 

\clearpage

\clearpage

%
%
\bibliographystyle{splncs04}
\bibliography{main}
\end{document}